\documentclass[conference]{IEEEtran}
\IEEEoverridecommandlockouts
\usepackage{cite}
\usepackage{amsmath,amssymb,amsfonts}
\usepackage{algorithmic}
\usepackage{booktabs}
\usepackage{graphicx}
\usepackage{textcomp}
\usepackage{xcolor}
\usepackage{listings}
\def\BibTeX{{\rm B\kern-.05em{\sc i\kern-.025em b}\kern-.08em
T\kern-.1667em\lower.7ex\hbox{E}\kern-.125emX}}
\begin{document}
%%%%%%%%%%%%%%%%%%%%%%%%%%%%%%
\title{AgentBnB: A Browser-Based Cybersecurity Tabletop Exercise with Large Language Model Support and Retrieval-Aligned Scaffolding}
%%%%%%%%%%%%%%%%%%%%%%%%%%%%%%
\author{
\IEEEauthorblockN{Arman Anwar}
\IEEEauthorblockA{
\textit{School of Electrical and Computer Engineering}\\
\textit{Georgia Institute of Technology}\\
Atlanta, USA\\
aanwar31@gatech.edu}
\and
\IEEEauthorblockN{Zefang Liu}
\IEEEauthorblockA{
\textit{School of Computational Science and Engineering}\\
\textit{Georgia Institute of Technology}\\
Atlanta, USA\\
liuzefang@gatech.edu}
}
%%%%%%%%%%%%%%%%%%%%%%%%%%%%%%
\maketitle
%%%%%%%%%%%%%%%%%%%%%%%%%%%%%%
\begin{abstract}
Traditional cybersecurity tabletop exercises (TTXs) provide valuable training but are often scripted, resource-intensive, and difficult to scale. We introduce AgentBnB, a browser-based re-imagining of the Backdoors \& Breaches game that integrates large language model teammates with a Bloom-aligned, retrieval-augmented copilot (C2D2). The system expands a curated corpus into factual, conceptual, procedural, and metacognitive snippets, delivering on-demand, cognitively targeted hints. Prompt-engineered agents employ a scaffolding ladder that gradually fades as learner confidence grows. In a solo-player pilot with four graduate students, participants reported greater intention to use the agent-based version compared to the physical card deck and viewed it as more scalable, though a ceiling effect emerged on a simple knowledge quiz. Despite limitations of small sample size, single-player focus, and narrow corpus, these early findings suggest that large language model augmented TTXs can provide lightweight, repeatable practice without the logistical burden of traditional exercises. Planned extensions include multi-player modes, telemetry-driven coaching, and comparative studies with larger cohorts.
\end{abstract}
%%%%%%%%%%%%%%%%%%%%%%%%%%%%%%
\begin{IEEEkeywords}
cybersecurity training, tabletop exercises, large language models, retrieval-augmented generation, instructional scaffolding
\end{IEEEkeywords}
%%%%%%%%%%%%%%%%%%%%%%%%%%%%%%
\section{Introduction}
\label{sec:introduction}

Cybersecurity tabletop exercises (TTXs)~\cite{angafor2020game,angafor2020bridging,young2021backdoors} have long served as a core method of incident-response training, giving teams structured environments to practice technical and organizational decision-making under pressure. Despite their widespread use in both industry and academia, conventional TTXs often fall short educationally. In classroom settings in particular, they can feel procedural, overly scripted, and logistically cumbersome, more like filling out a worksheet than responding to a live breach. Such rigidity limits scalability, reduces opportunities for iterative improvement, and constrains adaptation to individual learner needs.

Recent work has shown that large language model (LLM) agents and multi-agent systems can collaborate effectively, adapt to dynamic environments, and tackle complex tasks across diverse domains~\cite{li2023camel,park2023generative,wang2023voyager,liang2023encouraging,liu2024review,quan2024invagent,fan2024ai,guo2024large,wu2024autogen,zhang2024ai,liu2025econwebarena,quan2025crmagent}. Landmark studies such as \textit{Generative Agents}~\cite{park2023generative} and \textit{AutoGen}~\cite{wu2024autogen} highlight how LLMs can support emergent behavior and orchestrated multi-agent collaboration. Together, these works demonstrate the growing versatility of LLM-based agents and underscore their potential for structured problem solving. Building on this broader foundation, our work adapts these principles to the domain of cybersecurity training.

This project introduces \textbf{AgentBnB}, a lightweight, browser-based simulation platform that reimagines the TTX format through narrative gameplay, intelligent agents, and retrieval-augmented learning. Inspired by \textit{Backdoors \& Breaches}~\cite{backdoorsandbreaches}, a card game that teaches incident response, AgentBnB combines the game's procedural fidelity with large language model (LLM) teammates and an instructional copilot (C2D2\footnote{The name draws inspiration from R2D2 in Star Wars, but in this context it refers to a retrieval-augmented copilot designed to support decision-making and detection.}) that provides real-time, Bloom-aligned scaffolding.

Our work builds on AutoBnB's~\cite{liu2024multi,liu2025autobnb,liu2025autobnbrag} demonstration that LLM agents can autonomously play structured incident-response scenarios. Unlike AutoBnB's closed-loop AI simulation, which lacked human participation and pedagogical instrumentation, AgentBnB incorporates human-in-the-loop interaction, turn-based game-state management, and just-in-time instructional support. The result is a hybrid game and TTX experience that integrates simulation, collaboration, and reflective learning.

The key contributions of this work are:
\begin{enumerate}
    \item A browser-based interface with real-time chat and a stateful simulation engine that operationalizes the \textit{Backdoors \& Breaches} ruleset.
    \item An agent architecture enabling dynamic, role-based dialogue between human players and LLM teammates.
    \item A retrieval-augmented instructional copilot (C2D2) that delivers adaptive, Bloom-aligned guidance.
    \item A telemetry framework for capturing gameplay and copilot usage data to support research on usability and learning outcomes.
\end{enumerate}
Together, these elements provide learners with a lightweight way to practice incident response and reflect on their decisions without the logistical overhead of traditional tabletop exercises.

The remainder of this report is structured as follows. Section \ref{sec:problem-statement} examines the limitations of traditional TTXs that motivate our design. Section \ref{sec:related-work} reviews related work on tabletop exercises, game-based learning, and intelligent agents. Section \ref{sec:design-objectives} outlines the design objectives and requirements of AgentBnB. Section \ref{sec:system-overview} describes the system architecture, including C2D2's retrieval-augmented design. Section \ref{sec:evaluation-methodology} presents the evaluation methodology. Section \ref{sec:results} reports the results of our pilot study. Section \ref{sec:limitations} discusses the limitations of the current work, and Section \ref{sec:future-work} outlines directions for future research. Section \ref{sec:conclusion} summarizes the contributions and key findings.

%%%%%%%%%%%%%%%%%%%%%%%%%%%%%%
\section{Problem Statement}
\label{sec:problem-statement}

Although widely adopted, traditional cybersecurity tabletop exercises (TTXs) often fall short of their pedagogical potential. Evidence from prior research and classroom practice highlights five recurring limitations:
\begin{itemize}
    \item \textbf{Logistical overhead:} Effective exercises require extensive planning, cross-role coordination, and dedicated facilitation, which makes iteration impractical in many academic and training settings~\cite{young2021backdoors}.
    \item \textbf{Predictability:} Heavy reliance on predetermined scripts reduces realism and adaptability, diminishing the sense of urgency that characterizes real-world incidents~\cite{kick2014cyber}.
    \item \textbf{Role dilution:} Participants are frequently asked to assume multiple personas (e.g., technical lead, legal counsel, executive), which reduces authenticity and engagement~\cite{kilroy2024cyber}.
    \item \textbf{Lack of institutional memory:} Lessons learned are rarely formalized or integrated into curricula, weakening long-term educational impact~\cite{young2021backdoors}.
    \item \textbf{Underrepresentation of complexity:} Many exercises overlook interdependencies among stakeholders (e.g., legal, executive, and operations teams), failing to capture systemic failure modes common in high-stakes breaches~\cite{shreeve2020best,ahmad2020integration}.
\end{itemize}

These shortcomings motivate the central research question of this work:
\begin{quote}
\textit{Can we design a lightweight, repeatable, and immersive alternative to traditional TTXs that preserves procedural fidelity while improving learner engagement and knowledge retention?}
\end{quote}

%%%%%%%%%%%%%%%%%%%%%%%%%%%%%%
\section{Background and Related Work}
\label{sec:related-work}

A wide body of research has explored the design of cybersecurity tabletop exercises, educational games, and intelligent tutoring systems. This section situates AgentBnB within that landscape by reviewing traditional TTX practices, related game-based frameworks, and emerging applications of large language models (LLMs) in incident-response training.

%%%%%%%%%%%%%%%%%%%%%%%%%%%%%%
\subsection{Traditional Tabletop Exercises}

Tabletop exercises (TTXs) have long been used in cybersecurity education and readiness, providing structured role-play environments where teams rehearse incident-response workflows. These exercises typically simulate breaches using predefined scenarios, allowing participants to practice coordinated decision-making across technical, legal, and executive domains.

Despite their prevalence, traditional TTXs are often limited by rigidity. They emphasize procedural correctness over conceptual depth, restrict improvisation, and require extensive manual facilitation~\cite{young2021backdoors,kilroy2024cyber}. Such characteristics make them difficult to scale, iterate, or adapt to the needs of individual learners, particularly in academic or resource-constrained contexts.

Some dynamic variants~\cite{nespoli2024scorpion,gernhardt2025innovating,angafor2024malaware} have been proposed, including branching simulations and gamified adversary models, but these approaches often require costly infrastructure or intensive instructor involvement. These challenges highlight the need for lightweight, immersive, and repeatable alternatives that align more effectively with modern cybersecurity training goals.

%%%%%%%%%%%%%%%%%%%%%%%%%%%%%%
\subsection{Backdoors \& Breaches as a Training Framework}

AgentBnB builds on the mechanics and pedagogical goals of \textit{Backdoors \& Breaches} (B\&B)~\cite{backdoorsandbreaches}, a cybersecurity tabletop game developed by Black Hills Information Security. Our implementation extends B\&B into a digital simulation that supports cooperative play between human participants and AI teammates.

The core mechanics of the card game are preserved. Each session begins with a hidden sequence of four attack cards that represent phases of the adversary lifecycle: Initial Compromise, Pivot and Escalate, Persistence, and Command \& Control (C2) with Exfiltration. Players respond each turn by selecting from a set of Procedure cards, with outcomes determined by simulated dice rolls, cooldown logic, and randomly triggered inject events.

AgentBnB replaces physical materials with a browser-based conversational interface. The game state is managed entirely in memory, displayed to players through a minimal HUD, and logged for research purposes. By operationalizing the B\&B ruleset in this way, AgentBnB enables repeatable, scalable gameplay and creates opportunities for integration with intelligent agents and data-driven analysis.

%%%%%%%%%%%%%%%%%%%%%%%%%%%%%%
\subsection{Autonomous LLM Agents in Backdoors \& Breaches}

Recent work in AutoBnB~\cite{liu2025autobnb} has demonstrated the potential of LLMs as autonomous agents in structured incident-response simulations. AutoBnB deployed GPT-based agents to play \textit{Backdoors \& Breaches} without human participation, framing the game as a multi-agent coordination problem. The agents, organized into centralized, decentralized, and hybrid teams, achieved success rates of up to 36\% in uncovering complete attack chains without fine-tuning or retrieval augmentation.

While these results show that LLMs can play the game effectively, AutoBnB remained a closed-loop AI simulation. It did not support human interaction, lacked instructional scaffolding, and was not integrated with external knowledge sources.

AgentBnB builds on this foundation by introducing a hybrid architecture that combines human-in-the-loop interaction, persistent game state, and real-time instructional support through the C2D2 copilot. These extensions shift the focus from automation alone to pedagogy, transforming \textit{Backdoors \& Breaches} from a static simulation into an interactive learning environment.

%%%%%%%%%%%%%%%%%%%%%%%%%%%%%%
\subsection{Bloom's Taxonomy as a Lens for Cybersecurity Skill Development}

Bloom's revised taxonomy~\cite{krathwohl2002revision} provides a structured framework for assessing learning in cybersecurity tabletop exercises. It distinguishes progress along two dimensions: knowledge types (factual, conceptual, procedural, and metacognitive) and cognitive processes (remember, understand, apply, analyze, evaluate, and create).

AgentBnB applies this framework by expanding raw cybersecurity texts into discrete knowledge units aligned with Bloom's categories. Each snippet is stored in a vector database with its Bloom label, enabling the instructional copilot (C2D2) to retrieve knowledge at the cognitive depth most relevant to the learner.

During gameplay, C2D2 surfaces category-specific snippets that match the learner's query, allowing the LLM to ground its reasoning in appropriately scoped material: factual definitions for recall, conceptual models for understanding, procedural walkthroughs for application or analysis, and metacognitive heuristics for evaluation and creation.

This expansion-first strategy ensures that assistance is both pedagogically aligned and content-grounded, transforming AgentBnB from a simulation tool into a learning environment with integrated instructional support.

%%%%%%%%%%%%%%%%%%%%%%%%%%%%%%
\subsection{Instructional Scaffolding via Prompt Design}

Research on implicit scaffolding in interactive simulations~\cite{podolefsky2013implicit,van2010scaffolding} shows that carefully shaped prompts, affordances, and feedback can guide exploration without heavy-handed scripts. More recent work highlights how AI systems are reshaping educational ecosystems and professional learning~\cite{cukurova2024professional,niemi2024ai,mcdonald2025generative}, underscoring the importance of designing instructional support that is adaptive, transparent, and pedagogically aligned.

AgentBnB adopts this principle through prompt engineering. Each LLM teammate and the C2D2 copilot receives a system message that frames their primary mission as supporting the learner's growth. Prompts are applied dynamically using an eight-step progression, escalating support only when necessary and fading once mastery signals appear:
\begin{enumerate}
    \item Wait \& Observe: Allow independent reasoning and assess learner readiness.
    \item Prompt Self-Explanation: e.g., “Why did you select this procedure?” (Understand).
    \item Ask Targeted Questions: e.g., “What signals would indicate lateral movement?” (Analyze).
    \item Offer Analogies or Clues: e.g., “Think Equifax-style privilege escalation.” (Apply).
    \item Eliminate Red Herrings: e.g., “Could this alert be noise rather than signal?” (Evaluate).
    \item Narrow the Scope: e.g., “Focus your search on identity systems.” (Evaluate).
    \item Reveal Partial Solutions: e.g., “The attacker misused IAM roles.” (Apply).
    \item Reveal Full Solution: Provided only after objectives are met or at session end (Create).
\end{enumerate}

This prompt-driven approach keeps scaffolding lightweight and adaptive, ensuring that guidance is available when needed but unobtrusive once the learner demonstrates competence.

%%%%%%%%%%%%%%%%%%%%%%%%%%%%%%
\section{Design Objectives \& Requirements}
\label{sec:design-objectives}

This section translates the motivation established in previous sections into concrete objectives that guide implementation. It specifies the pedagogical goals, user-experience considerations, technical constraints, and research requirements that a viable implementation of AgentBnB must address.

%%%%%%%%%%%%%%%%%%%%%%%%%%%%%%
\subsection{Pedagogical Objectives}

Grounded in Bloom's revised taxonomy~\cite{krathwohl2002revision} and scaffolding theory~\cite{van2010scaffolding}, AgentBnB is designed to help learners progress across multiple levels of cognitive complexity. Table~\ref{tab:pedagogical-objectives} maps Bloom levels to the targeted competencies operationalized within the system.

\begin{table}[h]
\centering
\caption{Pedagogical objectives of AgentBnB aligned with Bloom's taxonomy}
\label{tab:pedagogical-objectives}
\begin{tabular}{ll}
\toprule
\textbf{Bloom Level} & \textbf{Targeted Competency in AgentBnB} \\
\midrule
Remember & Recall common incident-response terms and artifacts \\
Understand & Explain why a procedure (e.g., memory dump) is chosen \\
Apply & Execute appropriate procedures under time pressure \\
Analyze & Compare containment vs. eradication trade-offs \\
Evaluate & Critique effectiveness of actions post-incident \\
Create & Devise novel mitigation strategies \\
\bottomrule
\end{tabular}
\end{table}

Success is defined as measurable gains across at least three adjacent Bloom levels (e.g., Understand $\rightarrow$ Analyze) between pre- and post-surveys.

%%%%%%%%%%%%%%%%%%%%%%%%%%%%%%
\subsection{User‑Experience Goals}
\label{sec:ux-goals}

AgentBnB’s interface is intentionally streamlined to prioritize clarity, usability, and rapid development over production-grade polish. The design builds on three familiar paradigms that are widely recognized by both users and developers, ensuring immediate accessibility and minimizing cognitive overhead:
\begin{enumerate}
    \item \textbf{Group Chat (Game Channel):} the main pane where the learner and AI teammates conduct all incident-response dialogue.
    \item \textbf{Copilot Chat (C2D2 Channel):} a separate tab for retrieval-augmented tutoring, keeping instructional hints distinct from in-game conversation.
    \item \textbf{Compact Heads-Up Display (HUD):} a single status bar displaying turn number, dice outcomes, and remaining procedures.
\end{enumerate}

Restricting the interface to these elements minimizes cognitive load, reduces implementation overhead, and keeps the experimental focus on learning outcomes rather than interface novelty. Advanced features such as multi-window layouts or analytics dashboards are intentionally deferred to future iterations.

%%%%%%%%%%%%%%%%%%%%%%%%%%%%%%
\section{System Overview}
\label{sec:system-overview}

This section presents the architecture and operational flow of AgentBnB, a browser-based simulation platform for cybersecurity training. The system integrates four core components: a web-based interface, a structured \textit{Backdoors \& Breaches} game engine, multiple LLM-driven agents, and an instructional support module (C2D2) that delivers just-in-time feedback through a retrieval-augmented generation (RAG) pipeline. Together, these elements create an immersive, repeatable, and pedagogically aligned environment for conducting tabletop exercises.

%%%%%%%%%%%%%%%%%%%%%%%%%%%%%%
\subsection{Design Goals \& Architecture Principles}

\begin{figure}[h]
    \centering
    \includegraphics[width=\linewidth]{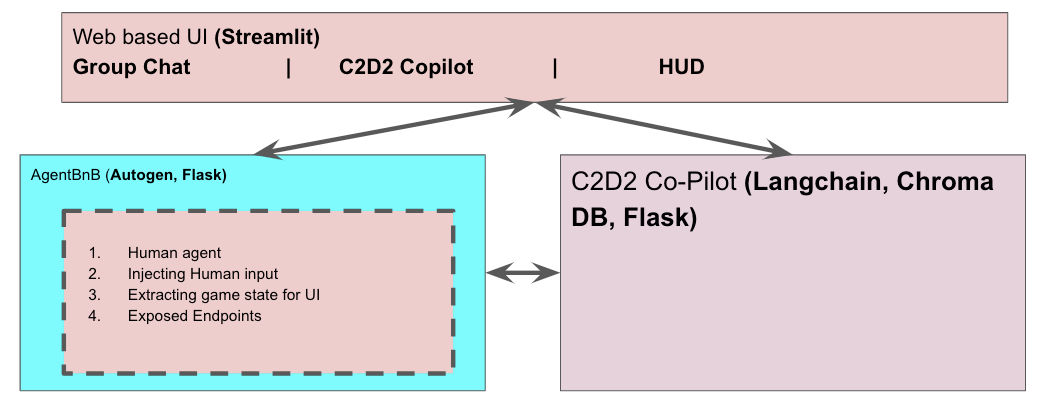}
    \caption{AgentBnB system architecture, showing the user interface, game engine, agent layer, C2D2 RAG module, and telemetry paths.}
    \label{fig:system-architecture}
\end{figure}

Figure~\ref{fig:system-architecture} (placeholder) illustrates the AgentBnB system architecture, including the user interface, game engine, agent layer, C2D2 RAG module, and telemetry paths.

AgentBnB is designed to support solo learners through a cooperative simulation that balances procedural realism with on-demand instructional support. The architecture is guided by three core principles:

\begin{enumerate}
    \item \textbf{Immersive Simulation:} gameplay should reflect the ambiguity and time pressure of real incident response, driven by a dynamic narrative.
    \item \textbf{Human-in-the-Loop:} one human participant collaborates with intelligent agent teammates in a shared decision-making environment.
    \item \textbf{Pedagogical Scaffolding:} instructional support must be timely, context-aware, and minimally disruptive to gameplay.
\end{enumerate}

%%%%%%%%%%%%%%%%%%%%%%%%%%%%%%
\subsection{Key Components}

AgentBnB is composed of several interdependent modules that together deliver the hybrid game-learning experience. Each component is designed to remain lightweight, modular, and easily extensible, allowing the system to support both research experimentation and future scaling.

%%%%%%%%%%%%%%%%%%%%%%%%%%%%%%
\subsubsection{Graphical User Interface (GUI)}

The AgentBnB interface adopts three familiar paradigms (see Section~\ref{sec:ux-goals}) to minimize learning overhead and maintain focus on gameplay and instruction:
\begin{enumerate}
    \item \textbf{Group Chat (Game Channel):} the central narrative thread where the learner, LLM teammates, and the Incident Master conduct all in-game dialogue.
    \item \textbf{Copilot Chat (C2D2 Channel):} a dedicated space for retrieval-augmented tutoring, providing Bloom-aligned hints and citations without disrupting the game flow.
    \item \textbf{Compact HUD (Bottom Bar):} a status bar that displays turn number, dice rolls, revealed attack cards, cooldown timers, and consecutive failures.
\end{enumerate}

%%%%%%%%%%%%%%%%%%%%%%%%%%%%%%
\subsubsection{Game Engine}

A lightweight, in-memory engine implements the \textit{Backdoors \& Breaches} ruleset, maintaining procedural fidelity while minimizing computational overhead:
\begin{enumerate}
    \item Four hidden Attack Cards (Initial Compromise through C2/Exfiltration) are randomly drawn at the start of each session.
    \item Procedure Cards include cooldown logic and gain bonuses when pre-documented (“written”).
    \item Each turn is resolved by a single d20 roll, with success defined as greater or equal to 11.
    \item Inject Cards are triggered by critical dice outcomes (natural 1, natural 20, or three consecutive failures).
\end{enumerate}
All state updates occur client-side, synchronizing the chat and HUD in real time without server round-trips.

%%%%%%%%%%%%%%%%%%%%%%%%%%%%%%
\subsubsection{Agent Layer}

\begin{figure}[h]
    \centering
    \includegraphics[width=\linewidth]{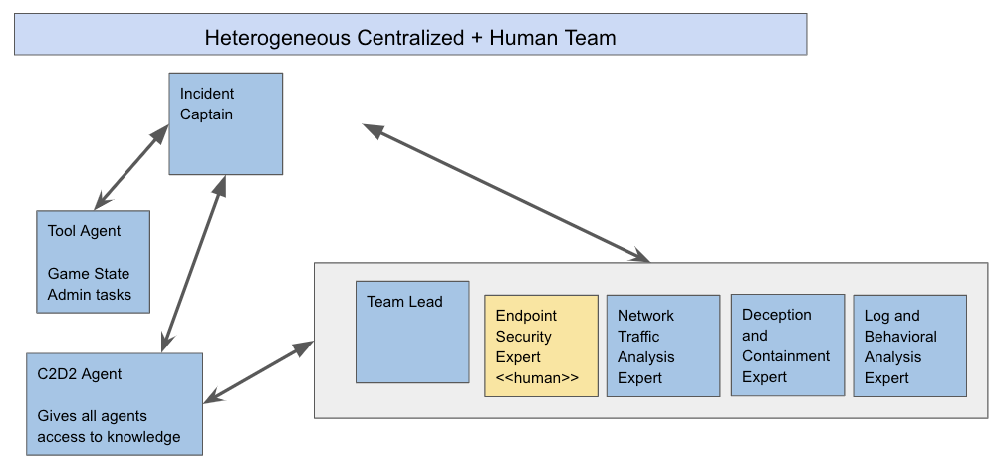}
    \caption{Social architecture of AgentBnB, including the human defender, Incident Captain, SOC Analyst, Red-Team Narrator, and their chat pathways.}
    \label{fig:social-architecture}
\end{figure}

Figure~\ref{fig:social-architecture} (placeholder) illustrates the social architecture of AgentBnB, including the human defender, Incident Captain, SOC Analyst, Red-Team Narrator, and their chat pathways. LLM agents serve dual roles as both cooperative teammates and instructional partners. Building on AutoBnB's~\cite{liu2025autobnb} fully autonomous architecture, we introduce several pedagogical enhancements:
\begin{enumerate}
    \item \textbf{Human Integration:} any LLM defender can be replaced by a live participant without disrupting team logic, enabling mixed human-AI play across AutoBnB's six organizational topologies.
    \item \textbf{Copilot Access for All Agents:} queries from both human and AI actors are routed through C2D2's RAG pipeline, ensuring a single authoritative knowledge source and reducing hallucination.
    \item \textbf{Pedagogical Prompt Augmentation:} system prompts prioritize teaching over winning and embed Bloom-aware scaffolding directives:  
    \begin{itemize}
        \item \textit{Teaching Objective:} “Your primary mission is to help the learner grow.”  
        \item \textit{Scaffolding Awareness:} apply contingency, fading, and transfer-of-responsibility strategies.
        \item \textit{Bloom Integration:} tailor support to the learner's current cognitive state.
    \end{itemize}
    \item \textbf{Incident Captain Prompt:} acts as a mentor who begins with observation, escalates through Socratic questioning and analogies, and provides direct guidance only when persistent misconceptions remain.
    \item \textbf{Defender Prompts:} encourage self-explanation and peer coaching, activating Bloom levels from \textit{Understand} (explain tools) to \textit{Create} (propose novel mitigations). An eight-level scaffolding rubric (prompting, probing, redirecting, hinting, etc.) governs the intensity of assistance.
\end{enumerate}
These enhancements reframe agents from performance-optimized bots into adaptive tutors, aligning \textit{Backdoors \& Breaches} with cognitive apprenticeship principles.

%%%%%%%%%%%%%%%%%%%%%%%%%%%%%%
\subsubsection{C2D2 Instructional Copilot}

The C2D2 module provides real-time instructional support through a retrieval-augmented generation (RAG) pipeline (see Section~\ref{sec:system-overview}). It surfaces Bloom-aligned knowledge snippets keyed to the learner's current cognitive level, with hints that gradually fade in specificity as mastery signals emerge. This design operationalizes adaptive scaffolding while reducing the risk of overreliance on system guidance.

%%%%%%%%%%%%%%%%%%%%%%%%%%%%%%
\subsubsection{Telemetry \& Logging}

Client-side telemetry hooks record all chat turns, dice rolls, and copilot queries, storing them in JSON Lines format within the browser. Export functionality to CSV or JSON enables post-hoc analysis without requiring persistent backend services, ensuring both lightweight deployment and research-grade observability.

%%%%%%%%%%%%%%%%%%%%%%%%%%%%%%
\subsection{Retrieval‑Augmented Generation (RAG) Architecture}

C2D2 employs a domain-adapted retrieval-augmented generation (RAG) system to provide grounded, context-sensitive instructional support. Compared to static, prompt-only approaches, the RAG pipeline offers three key advantages for cybersecurity education:

\begin{enumerate}
    \item \textbf{Factual grounding:} responses are anchored to authoritative sources, reducing hallucinations~\cite{lewis2020retrieval}.
    \item \textbf{Updatable knowledge:} new documents can be incorporated without model retraining~\cite{karpukhin2020dense}.
    \item \textbf{Context alignment:} retrieved passages are filtered and structured to match the learner's cognitive state, ensuring support is both accurate and pedagogically appropriate.
\end{enumerate}

\begin{figure}[h]
    \centering
    \includegraphics[width=\linewidth]{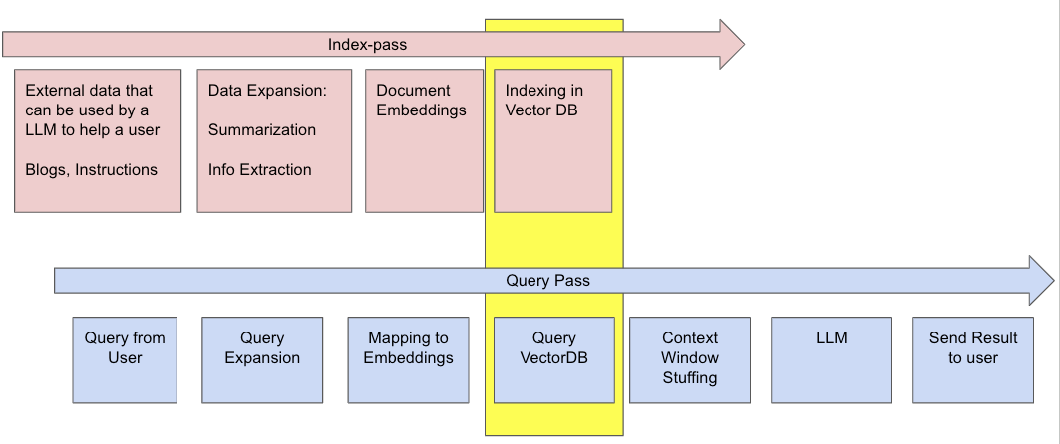}
    \caption{C2D2 two-stage flow, consisting of offline knowledge expansion and online retrieval and generation.}
    \label{fig:rag-architecture}
\end{figure}

Figure~\ref{fig:rag-architecture} (placeholder) illustrates C2D2's two-stage flow, consisting of offline knowledge expansion and online retrieval/generation.

%%%%%%%%%%%%%%%%%%%%%%%%%%%%%%
\subsubsection{RAG Overview}

Retrieval-augmented generation combines dense passage retrieval with autoregressive language modeling~\cite{izacard2020leveraging}. A learner query is embedded and matched against a pre-indexed vector database. The top-$k$ passages (default $k=10$) are inserted into a structured system prompt that also includes recent group-chat context and an instructional directive.

%%%%%%%%%%%%%%%%%%%%%%%%%%%%%%
\subsubsection{C2D2 Pipeline}

C2D2 implements a two-stage pipeline consisting of offline knowledge expansion and online retrieval/generation.

%%%%%%%%%%%%%%%%%%%%%%%%%%%%%%
\textbf{Offline Knowledge Expansion:}

\begin{enumerate}
    \item \textbf{Corpus Construction:} 77 publicly available web pages cited on \textit{Backdoors \& Breaches} cards (technical blogs and other online resources) were collected.
    \item \textbf{Bloom-Aligned Knowledge Extraction:} each document is processed four times using GPT-4 prompts tuned to Bloom categories:  
    \begin{enumerate}
        \item \textit{Factual:} discrete facts (e.g., “Mimikatz enables LSASS dump extraction”).
        \item \textit{Conceptual:} models or principles (e.g., “Privilege escalation widens the attacker's blast radius”).
        \item \textit{Procedural:} step-by-step methods.
        \item \textit{Metacognitive:} heuristics or reflection tips (e.g., “Prioritize logs with temporal correlation to alerts”).
    \end{enumerate}
    Example output snippets include:
    \begin{enumerate}
        \item \textit{Factual:} “deruke tools contains random scripts, tools, and techniques.”  
        \item \textit{Conceptual:} “Single-byte XOR encryption is a simple method of encrypting data by using a single byte …”  
        \item \textit{Procedural:} “1. Open the PowerShell script named ‘Single-Byte\_XOR.ps1' …”  
        \item \textit{Metacognitive:} “Regularly explore and experiment with random scripts and tools …”  
    \end{enumerate}
    \item \textbf{Pedagogical Justification:} categorization enables C2D2 to match responses to gameplay needs, such as factual recall for quick definitions, procedural snippets for in-game actions, and metacognitive advice for post-mortems.
    \item \textbf{Embedding:} each $\sim$300-token chunk is embedded using \texttt{text-embedding-ada-002}~\cite{brown2020language}, producing 1536-dimension vectors suitable for cosine-similarity search.
    \item \textbf{Vector Space Foundations:} dense embeddings extend classic vector-space and latent semantic indexing methods~\cite{salton1975vector,deerwester1990indexing} into high-dimensional semantic space.
    \item \textbf{Indexing in ChromaDB:} vectors are stored with HNSW indices plus metadata (\texttt{bloom\_tag, card\_id, name, type, description, tools, detection}, etc.), supporting sub-second retrieval during play.
\end{enumerate}

%%%%%%%%%%%%%%%%%%%%%%%%%%%%%%
\textbf{Online Retrieval \& Generation:}

\begin{enumerate}
    \item \textbf{Query Embedding:} learner inputs are embedded using the same \texttt{ada-002} model to ensure representational alignment.
    \item \textbf{Vector Search:} top-10 passages are retrieved via cosine similarity.
    \item \textbf{Contextual Prompt Assembly:} the response context combines (i) retrieved Bloom-labeled snippets, (ii) rolling group-chat history, and (iii) a pedagogical system prompt.
    \item \textbf{Response Generation:} GPT-4o~\cite{achiam2023gpt} generates a source-cited reply calibrated to the learner's cognitive stage.
\end{enumerate}

This architecture delivers responses that are accurate, adaptive, and aligned with Bloom's taxonomy while remaining fully browser-native. With these components, AgentBnB fulfills the lightweight, research-oriented requirements outlined in Section~\ref{sec:design-objectives}, while retaining sufficient structure for meaningful evaluation.

%%%%%%%%%%%%%%%%%%%%%%%%%%%%%%
\section{Evaluation Methodology}
\label{sec:evaluation-methodology}

This study employed a mixed-methods design to assess both the pedagogical impact and system-level dynamics of the LLM-enhanced AgentBnB simulation. Quantitative measures were used to track learning gains, while qualitative artifacts, including chat logs, free-text reflections, and copilot queries, captured how learners interacted with AI teammates and leveraged the C2D2 assistant.

%%%%%%%%%%%%%%%%%%%%%%%%%%%%%%
\subsection{Participants \& Procedure}

Four graduate-level volunteers were recruited from a local university network. Prior cybersecurity experience ranged from none ($n=2$) to introductory coursework ($n=2$). Each participant completed approximately five gameplay turns in a solo-play incident-response scenario.

The study followed a three-stage protocol:
\begin{enumerate}
    \item \textbf{Briefing:} a 10 minute overview of the \textit{Backdoors \& Breaches} rules and the research purpose.  
    \item \textbf{Gameplay Session:} participants completed a full scenario using AgentBnB (about 60 minutes), with telemetry logs collected throughout.  
    \item \textbf{Post-Survey:} participants completed an immediate post-questionnaire (Appendix \ref{sec:survey}) to capture perceptions and self-reported learning outcomes.  
\end{enumerate}

%%%%%%%%%%%%%%%%%%%%%%%%%%%%%%
\subsection{Instrument Design}

A bespoke survey instrument was developed to capture four dimensions of learner experience and outcomes. Table~\ref{tab:instrument} summarizes the sections, example items, and response formats. Attention checks (e.g., ``Select option 3 for this item'') were included to ensure response fidelity.

\begin{table}[h]
\centering
\caption{Survey instrument dimensions, sample items, and response formats}
\label{tab:instrument}
\begin{tabular}{p{2cm} p{3cm} p{2cm}}
\toprule
\textbf{Section} & \textbf{Example Item} & \textbf{Scale / Format} \\
\midrule
Baseline Gameplay Familiarity & ``How many physical B\&B sessions have you completed?'' & Numeric free-response \\
\midrule
Use Preferences \& Perceived Utility & ``I would use this agent-based version to prepare for a cybersecurity interview.'' & 5-point Likert (1 = Strongly Disagree, 5 = Strongly Agree) \\
\midrule
Comparative Effectiveness Judgments & ``Compared to the physical card game, this version was more scalable.'' & 5-point Likert \\
\midrule
Knowledge Assessment & ``Which phase best captures lateral movement?'' & Multiple-choice \\
\bottomrule
\end{tabular}
\end{table}

%%%%%%%%%%%%%%%%%%%%%%%%%%%%%%
\subsection{Data Collection \& Analysis}

Data were gathered from multiple sources to capture both learning outcomes and system interactions, enabling a mixed-methods evaluation of AgentBnB.

%%%%%%%%%%%%%%%%%%%%%%%%%%%%%%
\subsubsection{Quantitative}

Because this pilot captured post-session data only, analysis was limited to descriptive statistics. For each Likert item we report the mean, standard deviation, and range (see Table~\ref{tab:results}). A simple preference delta, calculated as Intention\_Agent minus Intention\_Card, was also computed for each respondent to gauge relative utility. Given the small sample size ($n=4$), these differences are summarized qualitatively rather than subjected to inferential testing. The single knowledge quiz (three items) yielded a ceiling effect, with all participants scoring 3/3. As a result, no additional statistical analyses were conducted.

%%%%%%%%%%%%%%%%%%%%%%%%%%%%%%
\subsubsection{Qualitative}

Open-response answers and chat transcripts were lightly coded to identify themes of cognitive engagement, terminology use, and reliance on C2D2. Copilot queries were further classified by Bloom level (Remember through Create) to characterize patterns of help-seeking behavior.

%%%%%%%%%%%%%%%%%%%%%%%%%%%%%%
\subsubsection{Telemetry}

Gameplay logs recorded turn duration, dice outcomes, hint frequency, and error streaks. In this pilot, these signals were used only as contextual indicators; in future studies they will support the development of an adaptive scaffolding model.

%%%%%%%%%%%%%%%%%%%%%%%%%%%%%%
\subsubsection{Attention Check Handling}

One participant failed the attention-check item. Given the exploratory nature of this pilot and the very small sample size, their data were retained but flagged. No outlier corrections were applied.

%%%%%%%%%%%%%%%%%%%%%%%%%%%%%%
\section{Results}
\label{sec:results}

Four graduate-level cybersecurity students ($n=4$) completed the post-session questionnaire. Descriptive statistics are reported in Table~\ref{tab:results} and summarized in the subsections below.

\begin{table}[h]
\centering
\caption{Post-session descriptive statistics ($n=4$)}
\label{tab:results}
\begin{tabular}{lccc}
\toprule
\textbf{Measure} & \textbf{n} & \textbf{Mean (SD)} & \textbf{Range} \\
\midrule
Prior physical B\&B runs (\#) & 4 & 7.25 (1.89) & 6--10 \\
Intention to use Agent version (1--5) & 4 & 4.25 (0.50) & 4--5 \\
Intention to use Card version (1--5) & 4 & 2.25 (0.50) & 2--3 \\
Knowledge score (0--3) & 4 & 3.00 (0.00) & 3--3 \\
\bottomrule
\end{tabular}
\end{table}

%%%%%%%%%%%%%%%%%%%%%%%%%%%%%%
\subsection{Baseline Familiarity}

Although participants did not have access to the physical card deck during this study, the post-survey asked them to estimate how many full runs of the physical game they had completed or could reasonably envision completing. Because these reports are retrospective and hypothetical rather than observed, they should be interpreted as a proxy for familiarity rather than a direct measure of hands-on experience.

%%%%%%%%%%%%%%%%%%%%%%%%%%%%%%
\subsection{Use Preferences \& Perceived Utility}

The agent-based system was rated positively ($M=4.25/5$). Willingness to rely solely on the physical deck was lower ($M=2.25/5$), suggesting a preference for the automated version when practicing individually.

%%%%%%%%%%%%%%%%%%%%%%%%%%%%%%
\subsection{Comparative Effectiveness Judgements}

Half of the participants (50\%) judged the agent version more effective, one participant (25\%) preferred the card game, and one participant (25\%) rated them as equal. Among those who favored the agent version, all rated the advantage as at least “Moderate.”

%%%%%%%%%%%%%%%%%%%%%%%%%%%%%%
\subsection{Knowledge Assessment}

All participants answered the three knowledge items correctly, producing a ceiling effect. Future studies will include more discriminating questions and adopt a pre/post design to better capture learning gains.

%%%%%%%%%%%%%%%%%%%%%%%%%%%%%%
\subsection{Study Constraints}

With only four participants (one failed the attention check) and no control condition, these results should be considered formative. Nonetheless, the directional preferences suggest value in expanding the study with a larger sample and more robust learning metrics.

%%%%%%%%%%%%%%%%%%%%%%%%%%%%%%
\section{Limitations}
\label{sec:limitations}

This prototype establishes the feasibility of blending LLM agents with Bloom-aligned retrieval in a browser-based tabletop exercise, but its scope is constrained: the current system only supports single-player play through a minimal interface and a narrow (about 70 post) knowledge corpus, limiting realism and immersion. The pilot study itself was underpowered ($n=4$), used single-item scales, and lacked a control condition, restricting claims about learning gains. Finally, despite retrieval grounding, LLM reliability remains a concern, as hallucinations or retrieval gaps can surface under ambiguous prompts.

%%%%%%%%%%%%%%%%%%%%%%%%%%%%%%
\section{Future Work}
\label{sec:future-work}

Several extensions could refine AgentBnB into a more scalable and immersive training platform:

\begin{itemize}
    \item \textbf{Richer dialogue realism:} current chat turns can feel mechanical. Adding humor, emotional tone, or structured disagreement (for example, Red Team push back) could improve engagement. Emerging multi agent coordination techniques~\cite{guo2024large} provide one promising direction.  
    \item \textbf{Enhanced interface:} planned upgrades include interactive card panels, turn tracking widgets, scenario graph visualizations, and replay analysis tools. These improvements aim to reduce cognitive load and increase replayability.  
    \item \textbf{Live threat intelligence feeds:} integrating regularly refreshed indicators of compromise (IOCs) would enable scenario generation that reflects real time threat landscapes, improving authenticity.  
    \item \textbf{Multi player scalability:} support for multiple human defenders, or hybrid human and machine teams, would enable studies of coordination, escalation paths, and organizational communication~\cite{kilroy2024cyber}.  
    \item \textbf{Telemetry driven coaching:} mining chat and copilot logs for patterns such as decision latency or repeated misconceptions could support automated after action reports and personalized remediation plans.  
    \item \textbf{Comparative experiments:} future studies will add conditions such as copilot versus no copilot, and expand recruitment across institutions to provide stronger causal evidence of learning gains.  
\end{itemize}

These directions align with the project's goal of building a scalable, data driven, and pedagogically rich cyber training environment.

%%%%%%%%%%%%%%%%%%%%%%%%%%%%%%
\section{Conclusion}
\label{sec:conclusion}

This work introduced AgentBnB, a lightweight, browser-based reimagining of the \textit{Backdoors \& Breaches} tabletop exercise that integrates LLM-driven agents and a retrieval-augmented instructional copilot. Through a small pilot with graduate students, the system demonstrated feasibility for delivering scalable, repeatable, and pedagogically aligned incident-response practice without the logistical overhead of traditional exercises. Although the study was limited by sample size, single player focus, and a narrow knowledge corpus, the results suggest that hybrid human and AI simulations can enrich cybersecurity training by combining procedural fidelity with adaptive scaffolding. Future research will extend AgentBnB to multi player modes, expand telemetry-driven feedback, and evaluate learning outcomes across larger and more diverse cohorts.

%%%%%%%%%%%%%%%%%%%%%%%%%%%%%%
\bibliographystyle{IEEEtran}
\bibliography{conference}
%%%%%%%%%%%%%%%%%%%%%%%%%%%%%%
\newpage
\appendix
%%%%%%%%%%%%%%%%%%%%%%%%%%%%%%
\subsection{Survey Instrument}
\label{sec:survey}

\subsubsection*{Section 1: Basics}

\begin{enumerate}
  \item Before today’s session, approximately how many full playthroughs (``runs'') of \textit{Backdoors \& Breaches} have you completed? \\
  \textit{(Required)}

  \item How long, in minutes, does one game take on Agent BnB? (Use your best estimate if unsure.)

  \item How long do you think a single game of the card-based \textit{Backdoors \& Breaches} would take (in minutes)?
\end{enumerate}

\subsubsection*{Section 2: The Big Interview}

It’s Tuesday, and you’ve just spoken with a recruiter about a software development position at a high-tech firm.  
The interview is scheduled for Friday.  

During the conversation, the recruiter mentions that the role involves building cybersecurity tools, and that the company is a leader in cybersecurity incident response---particularly in something called “Red Team Tabletop Exercises.”  

The job sounds exciting, and you’re eager to make a good impression---but you’ve never heard of Red Team Tabletop Exercises before.  

Later that day, a friend recommends a training resource called \textit{Backdoors \& Breaches}, which is used to simulate cyber incidents.  
Before rushing off, your friend shares two versions with you:
\begin{itemize}
  \item An agent-based interactive version, and
  \item A physical card-based version from his own collection.
\end{itemize}

Now, with just a few days to prepare, you’re deciding how to best use these resources to get up to speed.  

The agent-based version of \textit{Backdoors \& Breaches} is highly automated, with AI agents simulating the roles of other players.  
In contrast, the card-based version is designed for group play and requires a human facilitator.  

\begin{enumerate}
  \setcounter{enumi}{3}
  \item Do you think you could effectively use the card-based version on your own to prepare for the interview? \\
  Mark only one:
  \begin{itemize}
    \item Yes, easily
    \item Yes, but it would be difficult
    \item No, but I could still review the cards and instructions
    \item No, not at all
  \end{itemize}

  \item Use of Agent-Based Version: How likely would you be to use the agent-based \textit{Backdoors \& Breaches} system to prepare for your interview? \\
  Likert scale (1 = Very unlikely, 5 = Very likely)

  \item Use of Card-Based Version: How likely would you be to use only the card-based \textit{Backdoors \& Breaches} game to prepare? \\
  Likert scale (1 = Very unlikely, 5 = Very likely)

  \item Attention Check: Please select ``Neutral'' for the options below. \\
  Likert scale (1 = Very unlikely, 5 = Very likely)

  \item Which version (card or agent) would be more effective for helping you prepare? \\
  Mark only one:
  \begin{itemize}
    \item Card-based version more effective
    \item Both versions equally effective
    \item Agent-based version more effective
  \end{itemize}

  \item How much more effective? (if one was chosen above) \\
  Mark only one:
  \begin{itemize}
    \item Slightly
    \item Moderately
    \item Significantly
  \end{itemize}
\end{enumerate}

\subsubsection*{Section 3: Self-Assessed Knowledge}

\begin{enumerate}
  \setcounter{enumi}{9}
  \item Which stage comes first in the adversary lifecycle? \\
  Mark only one:
  \begin{itemize}
    \item Persistence
    \item Command \& Control
    \item Initial Compromise
    \item Exfiltration
  \end{itemize}

  \item ``Lateral Movement'' primarily refers to which? \\
  Mark only one:
  \begin{itemize}
    \item Escalating privileges on the same host
    \item Moving from one host to another within the network
    \item Exfiltrating data to an external server
    \item Establishing persistence
  \end{itemize}

  \item A common persistence mechanism is: \\
  Mark only one:
  \begin{itemize}
    \item Using stolen credentials to pivot
    \item Dropping a startup script or service
    \item Sending data out via DNS
    \item Capturing packets on the wire
  \end{itemize}
\end{enumerate}

%%%%%%%%%%%%%%%%%%%%%%%%%%%%%%
\subsection{Group Chat View of UI}
\label{sec:group-chat-ui}

\begin{figure}[h]
    \centering
    \includegraphics[width=.95\linewidth]{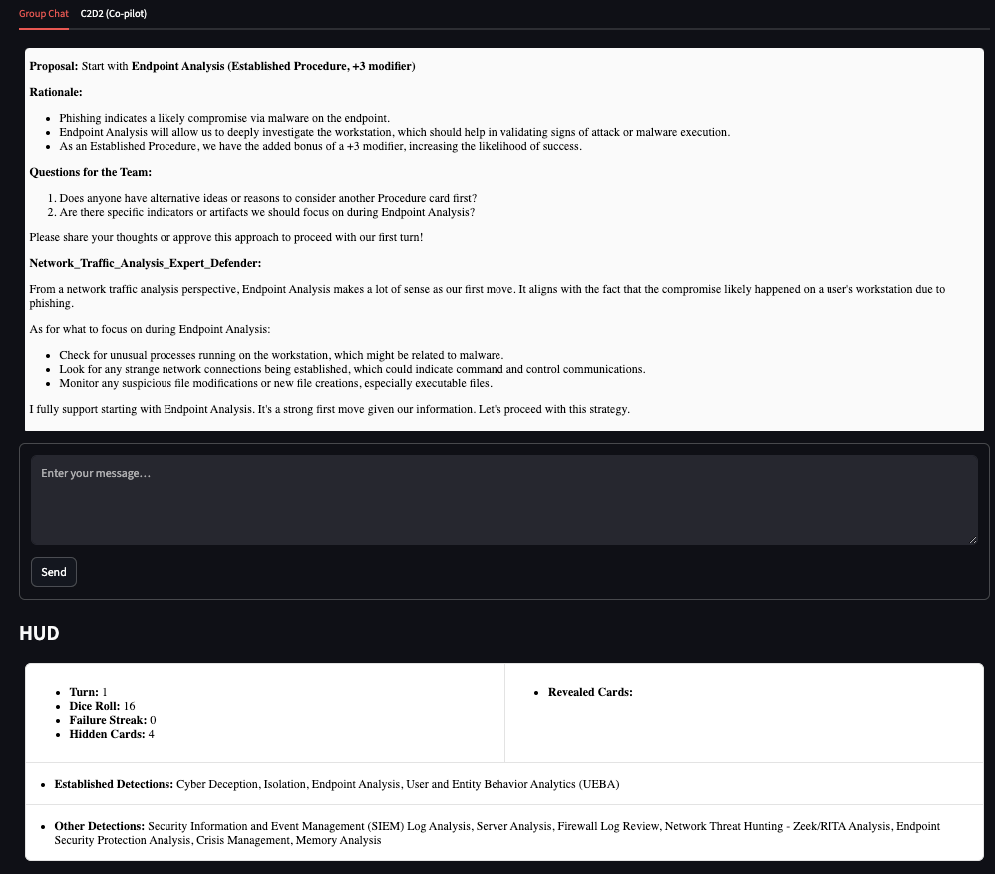}
    \caption{Group Chat view of the AgentBnB user interface, showing the main gameplay channel where the learner, AI teammates, and the Incident Master conduct in-game dialogue.}
    \label{fig:group-chat-ui}
\end{figure}

The Group Chat is the main pane for in-game dialogue (Figure~\ref{fig:group-chat-ui}).

%%%%%%%%%%%%%%%%%%%%%%%%%%%%%%
\subsection{C2D2 Chat View of UI}
\label{sec:c2d2-chat-ui}

\begin{figure}[h]
    \centering
    \includegraphics[width=.95\linewidth]{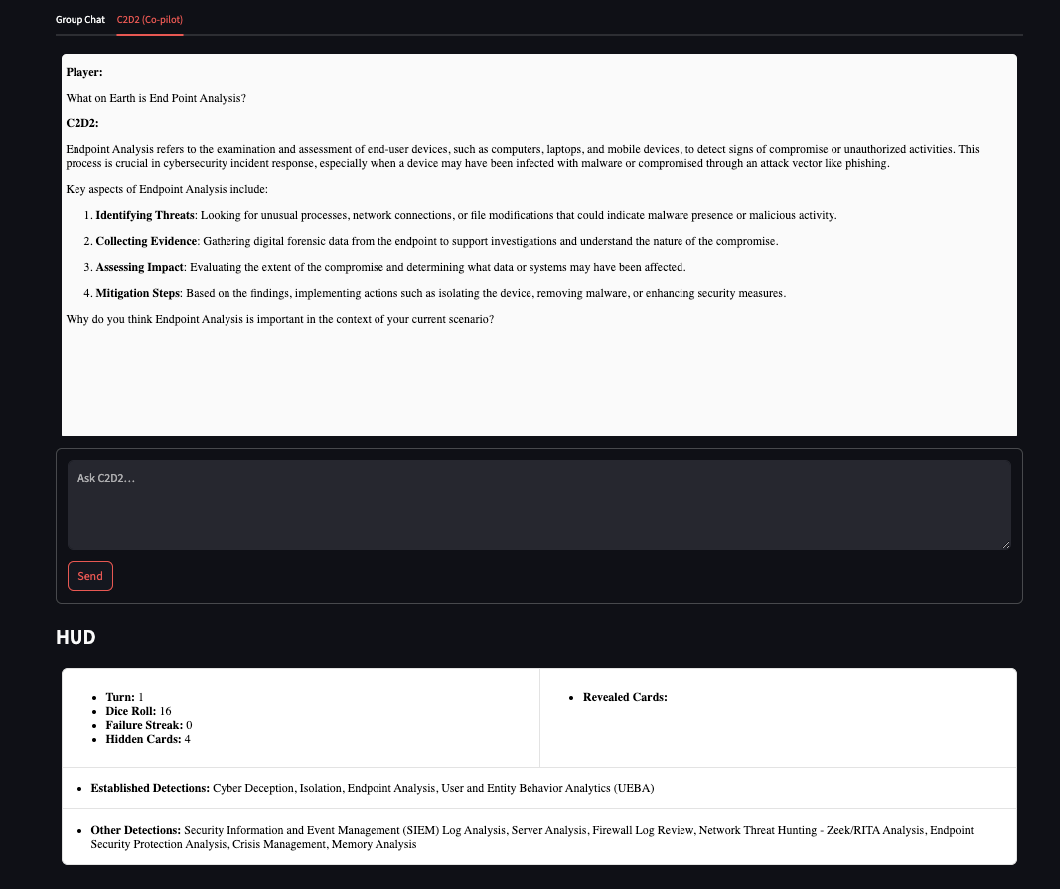}
    \caption{C2D2 Chat view of the AgentBnB user interface, showing the retrieval-augmented copilot channel that provides Bloom-aligned hints and citations separately from in-game dialogue.}
    \label{fig:c2d2-chat-ui}
\end{figure}

The C2D2 Chat provides retrieval-augmented hints and guidance (Figure~\ref{fig:c2d2-chat-ui}).

%%%%%%%%%%%%%%%%%%%%%%%%%%%%%%
\end{document}